\documentclass[sigconf]{acmart}

\usepackage{url}
\usepackage{subfigure}
\usepackage{color}
\usepackage{makecell}
\copyrightyear{2022}
\acmYear{2022}
\setcopyright{acmlicensed}
\acmConference[CIKM '22] {Proceedings of the 31st ACM International Conference on Information and Knowledge Management}{October 17--21, 2022}{Atlanta, GA, USA.}
\acmBooktitle{Proceedings of the 31st ACM International Conference on Information and Knowledge Management (CIKM '22), October 17--21, 2022, Atlanta, GA, USA}
\acmPrice{15.00}
\acmISBN{978-1-4503-9236-5/22/10}
\acmDOI{10.1145/3511808.3557168}


\definecolor{gray}{RGB}{150, 150, 150}
\definecolor{orange}{RGB}{255, 97, 0}
\definecolor{darkblue}{RGB}{25, 25, 112}

\def\sysname{o\underline{B}ject detection system for \underline{E}dge \underline{D}evices}
\def\sysnameb{Object Detection System for Edge Devices}
\def\sysnameabbr{BED}
\def\footnoteeqcon{*} 

\title{\sysnameabbr{}: A Real-Time Object Detection System for Edge Devices}

\begin{document}

\author{Guanchu Wang\footnoteeqcon{}}
\email{guanchu.wang@rice.edu}
\affiliation{%
  \institution{Rice University}
  \country{}
}

\author{Zaid Pervaiz Bhat\footnoteeqcon{}}
\email{zaid.bhat1234@tamu.edu}
\affiliation{%
  \institution{Texas A$\&$M University}
  \country{}
}

\author{Zhimeng Jiang\footnoteeqcon{}}
\email{zhimengj@tamu.edu}
\affiliation{%
  \institution{Texas A$\&$M University}
  \country{}
}

\author{Yi-Wei Chen\footnoteeqcon{}}
\email{yiwei\_chen@tamu.edu}
\affiliation{%
  \institution{Texas A$\&$M University}
  \country{}
}

\author{Daochen Zha\footnoteeqcon{}}
\email{daochen.zha@rice.edu}
\affiliation{%
  \institution{Rice University}
  \country{}
}

\author{Alfredo Costilla Reyes\footnoteeqcon{}}
\email{acostillar@rice.edu}
\affiliation{%
  \institution{Rice University}
  \country{}
}

\author{Afshin Niktash}
\email{Afshin.Niktash@analog.com}
\affiliation{%
  \institution{Analog Devices}
  \country{}
}

\author{Gorkem Ulkar}
\email{Gorkem.Ulkar@analog.com}
\affiliation{%
  \institution{Analog Devices}
  \country{}
}

\author{Erman Okman}
\email{Erman.Okman@analog.com}
\affiliation{%
  \institution{Analog Devices}
  \country{}
}

\author{Xuanting Cai}
\email{caixuanting@fb.com}
\affiliation{%
  \institution{Meta Platforms, Inc}
  \country{}
}

\author{Xia Hu}
\email{xia.hu@rice.com}
\affiliation{%
  \institution{Rice University}
  \country{}
}

\renewcommand{\shortauthors}{Guanchu Wang et al.}

% \author{
% Guanchu Wang*$^{1}$, 
% Zaid Pervaiz Bhat*$^{2}$, 
% Zhimeng Jiang*$^{2}$, 
% Yi-Wei Chen*$^{2}$, 
% Daochen Zha*$^{1}$, 
% Alfredo Costilla Reyes*$^{1}$, 
% Afshin Niktash$^{3}$, 
% Gorkem Ulkar$^{3}$, 
% Erman Okman$^{3}$, 
% Xuanting Cai$^{4}$, 
% Xia Hu$^{1}$
% }
% \thanks{* Those authors contribute equally to this project}
% \email{{guanchu.wang, daochen.zha, acostillar, xia.hu}@rice.edu, {zaid.bhat1234, zhimengj, yiwei\_chen }@tamu.edu}
% \email{
% {Afshin.Niktash, Gorkem.Ulkar, Erman.Okman}@maximintegrated.com, caixuanting@fb.com
% }
% \affiliation{%
%   \institution{$^{1}$Department of Computer Science, Rice University.}
%   \country{}
%   \institution{$^{2}$Department of Computer Science and Engineering, Texas A$\&$M University.}
%   \country{}
%   \institution{$^{3}$Maxim Integrated. $^{4}$Meta Platforms, Inc.}
%   \country{}
% %   \institution{$^{4}$Meta Platforms, Inc.}
%   \country{}
% }

\settopmatter{printacmref=true}
 

\begin{abstract}
% The combination of deep neural networks with edge computing shows significant potentials in the real-world application, especially for the scenarios that demand low latency, low power, or data privacy. 

% With the development of internet of thing technologies, 

Deploying deep neural networks~(DNNs) on edge devices provides efficient and effective solutions for the real-world tasks.
Edge devices have been used for collecting a large volume of data  efficiently in different domains.
DNNs have been an effective tool for data processing and analysis.
However, designing DNNs on edge devices is challenging due to the limited computational resources and memory. 
To tackle this challenge, we demonstrate \sysname{}~(\sysnameabbr{}) on the MAX78000 DNN accelerator.
It integrates on-device DNN inference with a camera and an LCD display for image acquisition and detection exhibition, respectively. 
\sysnameabbr{} is a concise, effective and detailed solution,
including model training, quantization, synthesis and deployment.
The entire repository is open-sourced on \href{https://github.com/datamllab/BED_main}{\color{darkblue} \texttt{Github}}\footnotemark, including a Graphical User Interface~(GUI) for on-chip debugging.
Experiment results indicate that \sysnameabbr{} can produce accurate detection with a 300-KB tiny DNN model, which takes only 91.9 ms of inference time and 1.845 mJ of energy.
The real-time detection is available at \href{https://youtu.be/0tY31_cECCA}{\color{darkblue} \texttt{YouTube}}\footnotemark.

% Our code is open-sourced on
% for the object detection on edge devices, 

% \centerline{\href{https://github.com/datamllab/BED_main}{{\color{darkblue} \texttt{https://github.com/datamllab/BED\_main}}}.}

% \noindent
% A demo video is available at

% \centerline{\href{https://www.youtube.com/watch?v=HZ9wwyOvjfI}{{\color{darkblue}\texttt{https://youtu.be/0tY31\_cECCA}}}.}

\end{abstract}

\maketitle

\newcommand\blfootnote[1]{%
\begingroup
\renewcommand\thefootnote{}\footnote{#1}%
\addtocounter{footnote}{-1}%
\endgroup
}
\blfootnote{* These authors contributed equally to this work.}
% \footnotetext{These authors contributed equally to this work.}

\vspace{-5mm}
\section{Introduction}

\footnotetext[1]{\small \href{https://youtu.be/0tY31_cECCA}{{\color{darkblue}\texttt{https://youtu.be/0tY31\_cECCA}}}}
\footnotetext[2]{\small \href{https://github.com/datamllab/BED_main}{{\color{darkblue} \texttt{https://github.com/datamllab/BED\_main}}}}

With the explosive growth of internet of thing technologies, billions of image data have been collected from the edge devices in different real-world scenarios~\cite{murshed2021machine}. For example, a single surveillance camera can collect nearly 60 GB traffic video per day~\cite{cctv-2007}. This enables us to leverage powerful deep neural networks~(DNNs) to process and analyze the data, which has various applications, such as action recognition, and object detection. However, DNN training/inference requires extensive computational resources, especially under the big data discipline. How we can efficiently deploy DNNs for image analysis on edges devices is an open challenge.

Traditional solutions embrace cloud-based services to transmit the collected data to the cloud for computation. Specifically, all the image data will be first uploaded to a cloud. Then a DNN will perform inference using powerful hardwares (e.g., GPUs) on the cloud. Finally, the inference results will be downloaded to the edge devices. However, such strategy suffers from high transmission delay~\cite{dillon2010cloud}.
To address this issue, edge computing has been proposed to transfer the major computations to devices~\cite{varghese2016challenges}. It brings the computation closer to the source of the data for fast and efficient data processing.
Moreover, it enjoys several benefits, such as low power consumption or risk of privacy invasion~\cite{shi2016edge}.

% the real-world application
% deploying and directly running machine learning (ML) models on edge devices have many important applications~\cite{murshed2021machine}. In contrast to the traditional cloud-based services that transmit the data to the cloud for computation, edge computing brings computation closer to the source of the data. As such, it enjoys several benefits, such as low latency, low power, and data privacy.

It is quite challenging to deploy DNNs to the edge devices due to the very limited hardware constraints, such as the memory capacity, computational resources or power consumption~\cite{wei2016security}.
\emph{First}, 
the common convolution operations usually needs high precision, e.g., 32-bit floating points, which consumes  massive computational resources (~\cite{devlin2018bert, dosovitskiy2020image, redmon2018yolov3}.
Edge devices support 16-bit floating points, 8-bit integer operations or less to simplify the hardware complexity~\cite{hassan2018role}.
% due to the limited computational resources.
\emph{Second}, DNNs require large memory to store the network weights and feature maps of intermediate layers during the feed-forward process~\cite{gao2020estimating}.
The memory capacity of edge devices is is not big enough for a large DNN model. 
Even though network pruning or distillation has been proposed to compress DNNs for the deployment~\cite{han2015deep}, it is difficult to maintain the performance of DNNs.
Therefore, deploying DNNs to edge devices requires non-trivial research and engineering efforts.

\begin{figure}
    \centering
    \setlength{\abovecaptionskip}{2mm}
    \setlength{\belowcaptionskip}{-2mm}
    \includegraphics[width=0.95\linewidth]{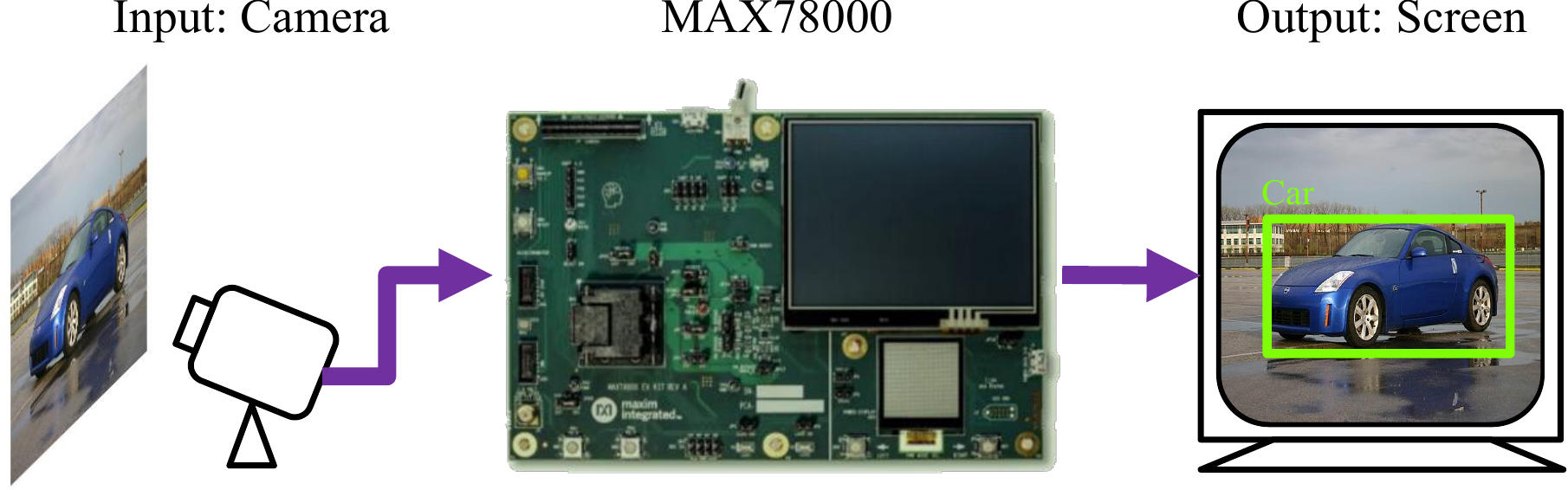}
    %\vspace{-8pt}
    \caption{\sysnameabbr{} implements a real-time and end-to-end object detection system from the camera to the screen.}
    %\vspace{-12pt}
    \label{fig:sys_config}
\end{figure}

%  especially deep neural networks (DNNs),
 
In this paper, we demonstrate the deployment of DNN models on edge devices for real-time object detection, which has broad real-world applications, such as surveillance, human computer interaction, and robotics~\cite{pathak2018application}. In particular, we present \sysname{}~(\sysnameabbr{}), an end-to-end system which integrates a DNN practiced on MAX78000 with I/O devices.
The system configuration is illustrated in Figure~\ref{fig:sys_config}.
Specifically, the DNN model for the detection is deployed on MAX78000, an efficient and low-power DNN accelerator; the I/O devices includes a camera and a screen for image acquisition and output exhibition, respectively.
The DNN model is pre-trained and evaluated on the VOC2007 dataset~\cite{pascal-voc-2007}.
Experiment results demonstrate \sysnameabbr{} can provide accurate object detection with a 300 KB tiny DNN model, and spend only 91.9 ms time and 1.845 mJ energy on the inference of each sample.
We also develop a Graphical User Interface~(GUI) in \sysnameabbr{} for users not familiar with the coding of MAX78000.
In the live and interactive part of our demo, we will showcase \sysnameabbr{} for real-time object detection.

\begin{figure}[t]
    \setlength{\abovecaptionskip}{2mm}
    \setlength{\belowcaptionskip}{-3mm}
    \centering
    \centering
    \includegraphics[width=0.99\linewidth]{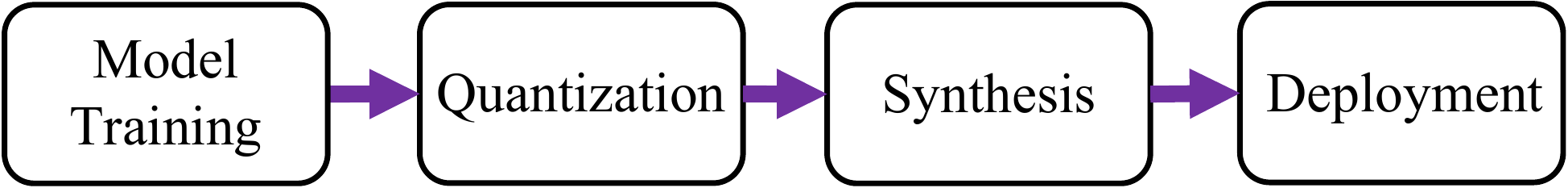}
    \caption{\label{fig:sys_pipeline}\sysnameabbr{} pipeline.}
    % \vspace{-5pt}
    % \caption{.}
\end{figure}

% To maximally compress the model, we develop a complete pipeline consisting of model training, quantization, synthesis, and deployment. \sysnameabbr{} can achieve competitive performance in terms of mean average precision (mAP) on a five-class object detection task compared with some popular TinyML models, such as Tiny YOLO-v1 and YOLO-v3 Tiny~\cite{adarsh2020yolo}, while only requiring 299.5KB memory cost. This is more than 100X smaller than the memory consumption of Tiny YOLO-v1 or YOLO-v3 Tiny.

%\noindent
%\textbf{Key points}:
%\begin{itemize}
%    \item An integrated system includes both processor and I/O devices.
%    \item The processor is an edge device far less powerful than CPU or GPU.
%\end{itemize}

% \vspace{-3mm}
\section{\sysnameb{}}

Figure~\ref{fig:sys_pipeline} shows the \sysnameabbr{} pipeline, which includes four stages: (i) model training stage that employs Quantization Aware Training to train a model, (ii) quantization stage that performs an 8-bit quantization, (iii) synthesis stage that converts the model to executable C code, and (iv) deployment stage that compiles the C code and loads the executable model to the edge device. We will first provide a background of MAX78000, and then elaborate on each of the stages.

% \vspace{-3mm}
\subsection{MAX78000 DNN Accelerator}
% {\color{red} TODO}

\begin{figure}
\setlength{\abovecaptionskip}{2mm}
\setlength{\belowcaptionskip}{-3mm}
\subfigure{
    \begin{minipage}[b]{0.37\linewidth}
    \includegraphics[width=1.0\linewidth]{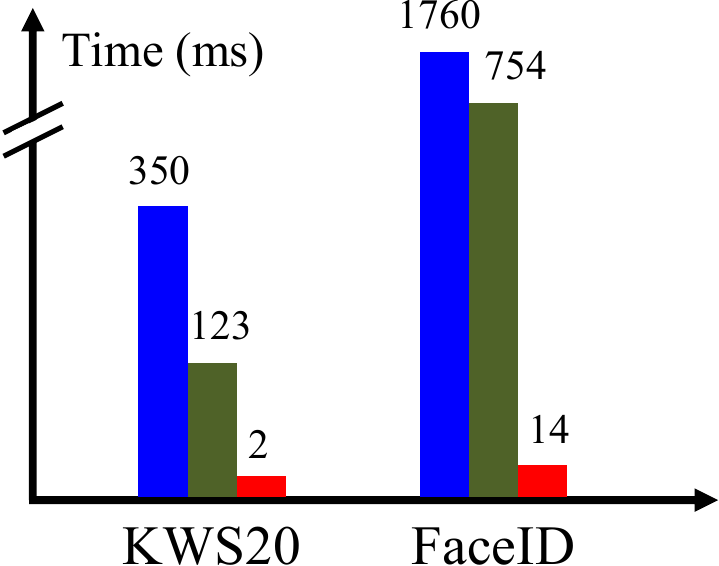}
    \end{minipage}
}
\subfigure{
    \begin{minipage}[b]{0.57\linewidth}
    \includegraphics[width=1.0\linewidth]{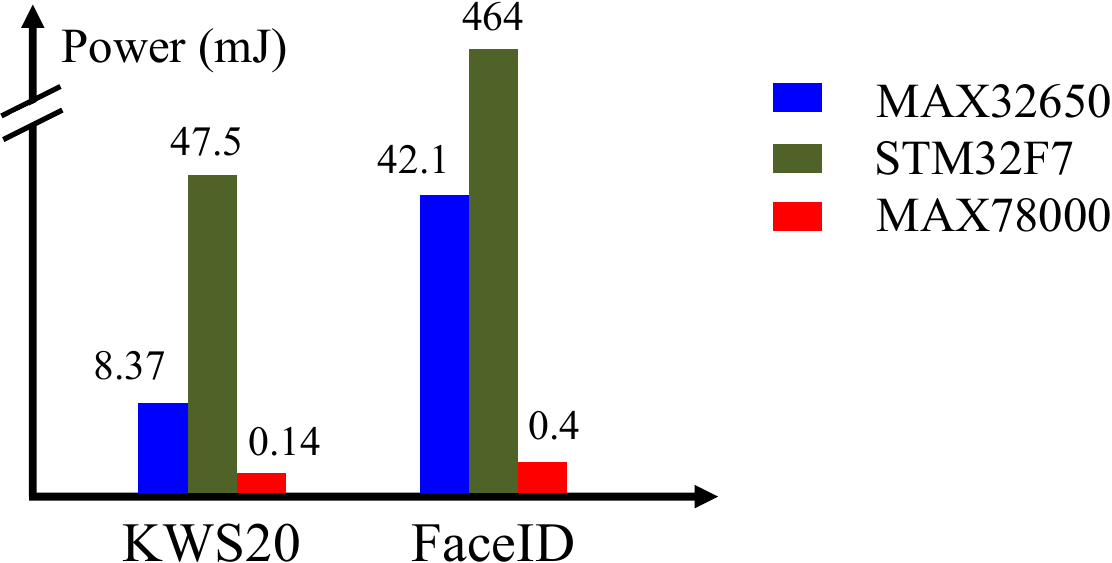}
    \end{minipage}
}
\caption{Comparison of inference time (left) and power consumption (right) of MAX78000 with two non-AI chips.\label{fig:time_power_comparison}}
\end{figure}

MAX78000 DNN Accelerator\footnote{ \url{https://www.maximintegrated.com/en/products/microcontrollers/MAX78000.html}} is a powerful AI microprocessor for efficient and low-power inference on edge devices.
Figure~\ref{fig:time_power_comparison} compares the inference time and the power consumption of MAX78000 with two non-AI microprocessors MAX32650\footnote{ARM DSP with CMSIS-NN, Cortex-M4 processor, working at 120 MHz.} and STM32F7\footnote{ARM DSP with CMSIS-NN, Cortex-M7 processor, working at 216 MHz.} on two representative DNN tasks KWS20\footnote{\scriptsize \url{https://github.com/MaximIntegratedAI/ai8x-training/blob/develop/models/ai85net-kws20.py}} and FaceID\footnote{\scriptsize \url{https://github.com/MaximIntegratedAI/ai8x-training/blob/develop/models/ai85net-faceid.py}}~\cite{Kristopher2020Cutting}. MAX78000 enjoys significant advantages in both inference time and power consumption. Thus, we deploy object detection models on MAX78000.

Despite its clear advantages, MAX78000 has several hard constraints on the model, making it challenging to design DNNs.
First, to speed up the inference, it only supports very few operators: 
3$\times$3 convolutional kernel, 1$\times$1 convolutional kernel, average pooling, maximum pooling, Relu activation function, etc.
Second, MAX78000 has only 432 KB flash for storing model parameters.
It is challenging to achieve a good performance under such operator and memory constraints.

%Hence, it is needed to adopt operator-level approximation for existing CNN backbones such as the GoogleNet~\cite{szegedy2015going}, MobileNet~\cite{howard2017mobilenets} or EfficientNet~\cite{tan2019efficientnet}.

% \begin{table}[h]\small
% \centering
% \begin{tabular}{l|l|l|l}
% \hline
%     Model & MAX32650 & STM32F7 & MAX78000\\
% \hline
%     KWS20   &    350ms, 8.37mJ	&	123ms, 47.5mJ  &	\bf{2.0ms, 0.14mJ}    \\
% \hline
%     FaceID  &   1760ms, 42.1mJ &   754ms, 464mJ   &	\bf{13.89ms, 0.40mJ}  \\  
% \hline
% \end{tabular}
% \caption{Inference time and energy consumption.}
% \label{tab:time_energy}
% \end{table}

% MACC\footnote{Multiply and accumulate operations} & 
% &	13801088  
% &	55234560  

\begin{figure*}
    \setlength{\abovecaptionskip}{2mm}
    \setlength{\belowcaptionskip}{-3mm}
    \centering
    \begin{minipage}[b]{0.8\linewidth}
    \includegraphics[width=1.0\textwidth]{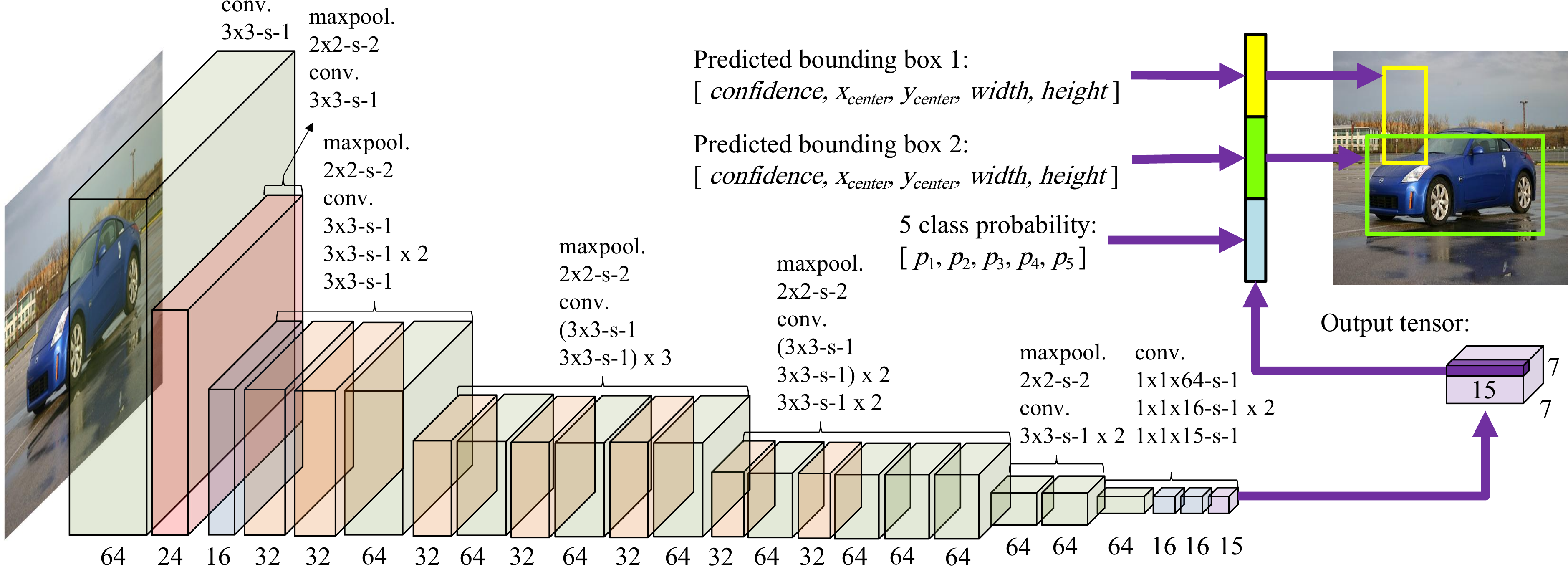}
    \caption{\label{fig:model_structure}DNN model structure.}
    \end{minipage}
    % \begin{minipage}[b]{0.35\linewidth}
    % \includegraphics[width=1.0\linewidth]{gui.pdf}
    % \caption{\label{fig:gui}GUI.}
    % \end{minipage}
\end{figure*}

% \vspace{-2mm}
\subsection{Model Training}
This subsection introduces the neural architecture of \sysnameabbr{} and the training details. We focus on standard object detection tasks, which aim to learn a DNN to detect the coordinate and the class of the existing objects from an image.
%We design and train the CNN model for the \sysnameabbr{} in this section.

%Specifically, \sysnameabbr{} follows the general object detection to learn a CNN model to detect the coordinate and class of the existing objects from an input image.
To maximally utilize the limited memory for model parameters, \sysnameabbr{} adopts a fully convolutional networks~\cite{long2015fully} for the detection.
Note that MAX78000 supports very limited operators.
The DNN model is constructed fully based on 3$\times$3 convoluational layer, Relu activation, Batch normalization, 2$\times$2 max-pooling without other operators, as shown in Figure~\ref{fig:model_structure}.
In this way, \sysnameabbr{} spends only 300KB on the storage of model parameters.
An input image is in the RGB format with a size of 224$\times$224$\times$3, and is divided into a 7$\times$7 grid, where each cell has a 32$\times$32 area.
The model outputs a 7$\times$7$\times$15 tensor for each image, where each of the 7$\times$7 cells corresponds to a 15 dimensional output vector consisting of class probabilities, two bounding boxes and their confidence scores.
In such a manner, the model outputs the detection result which contains both the coordinates and the class.

% including ten dimensional two predicted bounding boxes and five dimensional class prediction, where each bounding box five dimensional vector 
% the confidence of prediction, the two dimensional coordinate of center and the width and height of the bounding box.
% : person, car, aeroplane, cat and dog

The model is trained on a subset of VOC2007 dataset\footnote{\url{http://host.robots.ox.ac.uk/pascal/VOC/voc2007/index.html}} which contains the images of five classes with their annotations.
To minimize the performance degradation after the post-quantization, we adopt Quantization Aware Training~\cite{krishnamoorthi2018quantizing}.
Specifically, in the training, the model has the feed-forward process given by
\begin{equation}
    \mathbf{H}_{l+1} = \mathrm{Q} \big[ f(\mathbf{W}_l \mathbf{H}_l + \mathbf{b}_l) \big],
\end{equation}
where $\mathbf{H}_l, \mathbf{W}_l$ and $\mathbf{b}_l$ denote the feature map, weights and bias of layer $l$, respectively; $\mathrm{Q}(\bullet)$ denotes a simulative 8-bit quantization adopting FLOAT32 to simulate the INT8 inference.
We follow the training strategy of Yolo-V1~\cite{redmon2016you} to update the model. Specifically, we adopt cross entropy and mean square error loss functions for the classification and the bounding box coordinates, respectively.
Furthermore, we employ the SGD optimizer with $3 \times 10^{-5}$ learning rate, and adopt the mini-batch updating with batch-size 16 to update the parameters of model for 400 epochs;
we back up the snapshot of model at the end of each training epoch; and select the optimal model to maximize the mean average precision on the validating dataset.
More training details are provided in our repository\footnote{\url{https://github.com/datamllab/BED_main}}.

% \footnote{\scriptsize \url{https://github.com/MaximIntegratedAI/ai8x-training}}.

% The whole training process is based on the Maxim QAT training library

% Each Batch-normal layer is merged to its front convolutional layer during the training for deploying the model after the training.
% During the training process, each Batch-normal layer is merged to its front convolutional layer for the deployment to the hardware.

\begin{figure}
    \setlength{\abovecaptionskip}{2mm}
    \setlength{\belowcaptionskip}{-3mm}
    \includegraphics[width=0.8\linewidth]{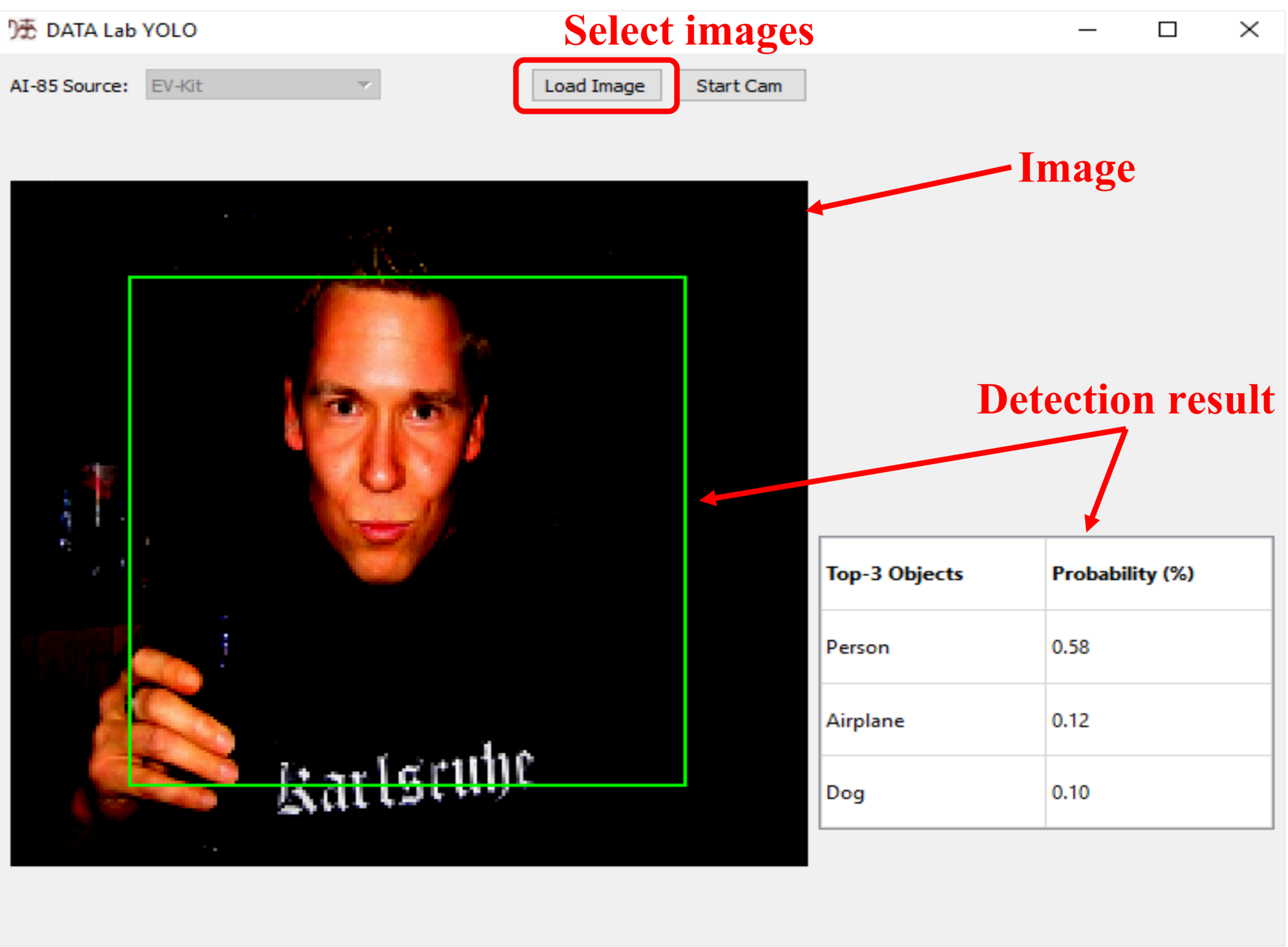}
    \caption{\label{fig:gui} GUI for BED.}
\end{figure}

% \vspace{-2mm}
\subsection{Quantization and Synthesis}

The trained model will be processed by quantization and synthesis such that it can be deployed on MAX78000.

The quantization stage reads a checkpoint file of the float pre-trained model and outputs the corresponding quantized model. 
During this process, the pre-trained model is processed by a 32-bit quantization for the last layer and an 8-bit quantization for the remaining layers, which involves the quantization of weight, bias and activation function for each layer of the model.
The code of quantization is available in our repository\footnote{ \url{https://github.com/MaximIntegratedAI/ai8x-synthesis/blob/develop/quantize.py}}.

The synthesis stage converts the quantized pre-trained model into C program.
Specifically, it reads the checkpoint of the pretrained model after the quantization and automatically generate header files to store its weights, bias and hyper-parameters. It also wraps up other requirements including the configuration files for the deployment.

% Our synthesis code is available here\footnote{\scriptsize \url{https://github.com/MaximIntegratedAI/ai8x-synthesis}}.

% \sysnameabbr{} adopts MAX78000\footnote{\scriptsize \url{https://github.com/MaximIntegratedAI/MAX78000_SDK}} CNN accelerator for the deployment of model following the pipeline in Figure~\ref{fig:synthsis_pipeline}.

\begin{figure*}[h]
\setlength{\abovecaptionskip}{2mm}
\setlength{\belowcaptionskip}{-3mm}
\centering
\subfigure[Offline detection results.]{
\begin{minipage}[b]{0.99\linewidth}
\centering
\includegraphics[width=1.0\linewidth]{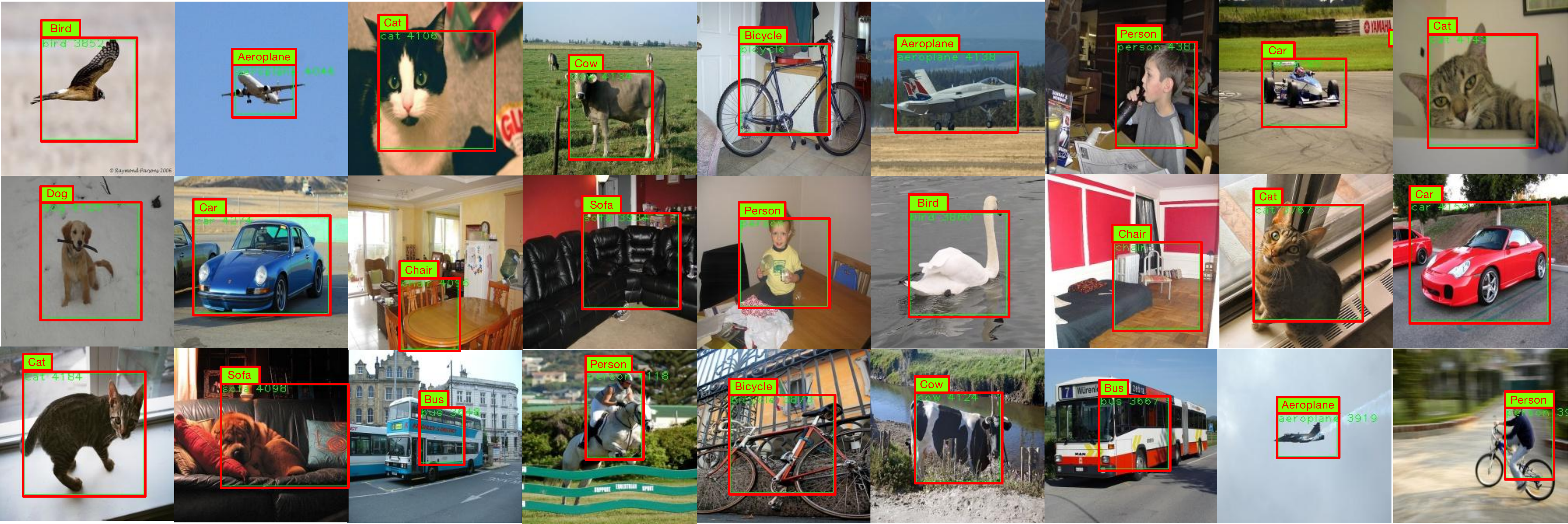}
% \caption{\label{fig:offline_result}Offline results.}
\end{minipage}
}
\subfigure[Testing bed.]{
    \begin{minipage}[b]{0.23\linewidth}
    \includegraphics[width=1.0\linewidth]{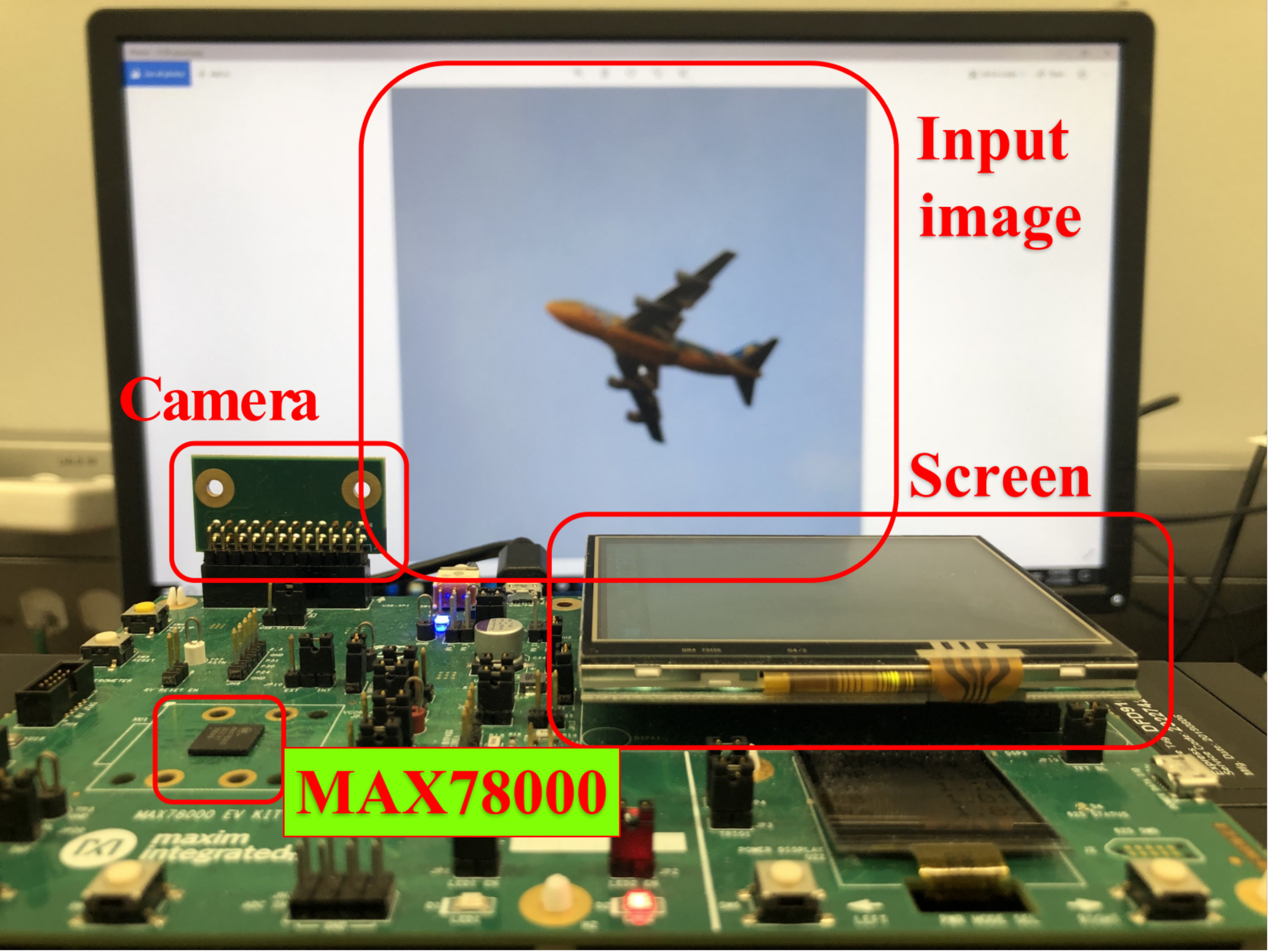}
    \end{minipage}
}
\subfigure[Real-time detection results.]{
    \begin{minipage}[b]{0.74\linewidth}
    \includegraphics[width=1.0\linewidth]{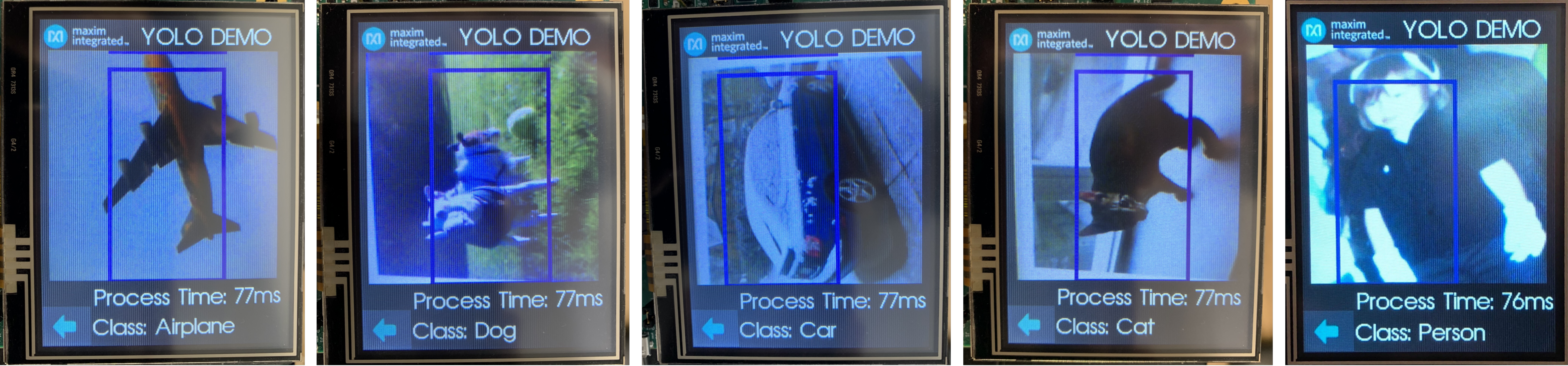}
    \end{minipage}
}
\caption{\label{fig:real_time_exp} Evaluation and demonstration.}
\end{figure*}

% \vspace{-2mm}
\section{Deployment}

After the synthesis, the C program is compiled into executable code using the ARM embedding compiler\footnote{{\fontsize{4.6}{4.6}\selectfont \url{https://developer.arm.com/tools-and-software/open-source-software/developer-tools/gnu-toolchain/gnu-rm/downloads.}}}.
The executable model is loaded to MAX78000 through serial protocol.
The source code of the deployment is given here\footnote{ \url{https://github.com/datamllab/BED_GUI}}\footnote{ \url{https://github.com/datamllab/BED_camera}}.

As an integrated system, \sysnameabbr{} adopts a camera\footnote{ \url{https://www.ovt.com/wp-content/uploads/2022/01/OVM7692-PB-v1.7-WEB.pdf}} to capture the images and an LCD screen\footnote{{\fontsize{5}{5}\selectfont \url{https://www.crystalfontz.com/products/document/3032/CFAF320240F-035T-TS_Data_Sheet_2012-04-11.pdf}}} to display the detection results.
The image captured by the camera is represented as a three-dimensional matrix with three channels of red, green and blue.
The images are loaded to MAX78000 block by block, where the blocks are temporarily stored to a flash memory with 896KB capacity inside MAX78000 until all of the blocks have been loaded.
After this, the model takes as input the image and outputs a 7$\times$7$\times$15 tensor to indicate the classification and bounding boxes in each grid of the input image.
The global result is selected from the grid-level 7$\times$7$\times$15 tensor via Non-maximum Suppression~\cite{redmon2016you}, where the global result is a $15$-dimensional vector to indicate the detection result for the whole input image, as shown in Figure~\ref{fig:model_structure}.
Finally, the result of classification and bounding box will be displayed on the LCD screen.

% \begin{table}[]
% \setlength{\abovecaptionskip}{2mm}
% \setlength{\belowcaptionskip}{-4mm}
%     \centering
%     \begin{tabular}{c|c}
%         \hline
%         Device & Value \\
%         \hline
%         Camera &  \\
%         Screen & \\
%         \hline
%     \end{tabular}
%     \caption{Caption}
%     \label{tab:my_label}
% \end{table}

% \vspace{-2mm}
\section{Evaluation and Demonstration}

% \vspace{0mm}
\subsection{Offline Evaluation}

The offline experiment focuses on evaluating the performance of the pre-trained model~(after quantization) before loading it to the edge device.
We visualize the detection results of some randomly chosen images from the testing set of VOC2007 in Figure~\ref{fig:real_time_exp}~(b). 
It is observed \sysnameabbr{} can accurately detect the objects in the input images.

% Specifically, we measure the mean average precision~(mAP) of the pre-trained model on the VOC2007 testing set, and report the evaluation results in Table~\ref{tab:mAP_table}, where we compare our \sysnameabbr{} with Yolo-v3 tiny\footnote{\url{https://github.com/ultralytics/yolov3}} and Tiny yolo\footnote{\url{https://pjreddie.com/darknet/yolo/}}.
% It is observed that \sysnameabbr{} achieves competitive performance with more than 100X less memory-cost and much smaller bit-width.

% \begin{table}[h]\small
% \centering
% \begin{tabular}{l|c|c|c|c}
% \hline
%     Model & mAP & Mem-cost & Bits & Device \\
% \hline
%     Yolo-v3 tiny   &	33.1   &	35.2MB	&	32 &   GPU \\
% \hline
%     Tiny yolo-v1  &	23.7    &	42.87MB &   32  &	GPU \\  
% \hline
%     Ours     &	18.96   &	299.5KB &	8   &	MAX78000 \\
% \hline
% \end{tabular}
% \caption{Offline evaluation results.}
% \label{tab:mAP_table}
% \end{table}

% \vspace{-2mm}
\subsection{Real-time Demonstration}

% \begin{figure}
%     \centering
%     \includegraphics[width=0.5\linewidth]{testing-bed-note.pdf}
%     \caption{Caption}
%     \label{fig:my_label}
% \end{figure}

We conduct two real-time experiments to demonstrate the on-device object detection.
The first experiment focus on on-device inference.
Specifically, a computer will transmit images to MAX78000, which then sends the detection results back to the computer.
In the computer-side, we implement a Graphical User Interface~(GUI) to send the image, receive and show the detection results, as shown in Figure~\ref{fig:gui}.
For more results, please refer to our~\href{https://youtu.be/0tY31_cECCA}{\color{darkblue} demo video}\footnote{\small \href{https://youtu.be/0tY31_cECCA}{{\color{darkblue}\texttt{https://youtu.be/0tY31\_cECCA}}}}
%More results of are given in our demonstration video.

The second experiment demonstrates the whole pipeline of \sysnameabbr{} based on the testing bed in Figure~\ref{fig:real_time_exp}~(b), where an image is shown on a source screen; \sysnameabbr{} captures the image by the camera and shows the detection results on the screen.
We randomly select some images from the testing set of VOC2007, and give the real-time detection results in Figure~\ref{fig:real_time_exp}~(c).
\sysnameabbr{} can correctly detect the object in the image captured by the camera and show the results on the screen.

\subsection{Latency of Real-time Detection}

The latency of \sysnameabbr{} is measured by averaging $100$ times of on-chip inference, which starts from an image loading to the output of detection results.
The average inference time and energy of \sysnameabbr{} are given in Table~\ref{tab:power}, which are 91.9 ms and 
1.845 mJ, respectively.
Moreover, \sysnameabbr{} merely requires 299.52 KB memory to store the network wegiths of the deep object detection model.
Generally, the latency of \sysnameabbr{} satisfies the demands of real-world scenarios in terms of the constraints of latency, energy and memory.

\begin{table}[h]
    \centering
    \setlength{\abovecaptionskip}{1mm}
    \setlength{\belowcaptionskip}{-3mm}
    \caption{\label{tab:power} Latency of BED.}
    \begin{tabular}{c|c|c|c}
    \hline
        Processing & Power & Inference time & Energy \\
    \hline
        \makecell[c]{Image loading \\ and DNN inference} & 20.08 mW & 91.9 ms & 1.845 mJ \\
    \hline
    \end{tabular}
\end{table}

% \noindent
% \textbf{Key points}:
% \begin{itemize}
%     \item Camera-board-screen pipeline (picture)
%     \item Bounding box illustration
%     \item On-board inference time
%     \item GUI tool (optional) 
%     \item Bounding box illustration on the GUI (optional) 
% \end{itemize} 
% \noindent
% \textbf{2 figures}:
% \begin{itemize}
%     \item Picture of physical testing bed
%     \item on-board screen illustration. 
% \end{itemize} 

% \vspace{-2mm}
\section{Live and Interactive Parts}

In the demo session, we will present a live demo of real-time object detection based on MAX78000 using our developed GUI. 
Moreover, we will give a tutorial of \sysnameabbr{}, including platform setup, model deployment and application of our developed GUI. 

% \vspace{-2mm}
\section{Conclusion and Future Work}

In this work, we build an integrated system, called \sysnameabbr{}, for real-time object detection on edge devices. 
We design a compact deep learning model under very limited memory, energy and computational resources.
\sysnameabbr{} captures images with a camera, computes on-chip inference of the DNN, and displays the detection results with an LCD screen.
% \sysnameabbr{} shows the promise in deploying complex DNNs to the edge devices . 
In the future, we will explore neural architect search to optimize network architectures under the constrains of memory, latency and power.

%% The file named.bst is a bibliography style file for BibTeX 0.99c
% \bibliographystyle{named}
\bibliographystyle{ACM-Reference-Format}
\balance
\bibliography{reference}

% \newpage
% \noindent
% \textbf{2.1 Key points}:
% \begin{itemize}
%     \item Model structure
%     \item Template of output vector
%     \item QAT Training
%     \item BN layer fusion
%     \item Post Quantization
% \end{itemize}

% \noindent
% \textbf{2.3 Key points}:
% \begin{itemize}
%     \item VOC2007 dataset\footnote{\url{http://host.robots.ox.ac.uk/pascal/VOC/voc2007/index.html}}
%     \item mAP results
%     \item Bounding box illustration
% \end{itemize}

% \noindent
% \textbf{1 figure}:
% \begin{itemize}
%     \item Bounding box illustration 
% \end{itemize} 

% \noindent
% \textbf{1 table}:
% \begin{itemize}
%     \item mAP 
% \end{itemize} 

\end{document}